\documentclass{article}
\usepackage{spconf,amsmath,graphicx, amssymb}
\usepackage{url}
\usepackage{subcaption, caption}
\usepackage[svgnames]{xcolor}
\usepackage{framed}
\definecolor{shadecolor}{named}{LightGray}

\title{Can local vision-language models improve activity recognition over vision transformers? - case study on newborn resuscitation}
\name{  
Enrico Guerriero, Kjersti Engan, Øyvind Meinich-Bache
}
\address{
University of Stavanger, Dept. of electrical eng. and computer science \\
Stavanger, Norway \\  \{kjersti.engan, oyvind.meinich-bache\}@uis.no 
}
\begin{document}

\maketitle
\begin{abstract}
Accurate documentation of newborn resuscitation is essential for quality improvement and adherence to clinical guidelines, yet remains underutilized in practice. Previous work using 3D-CNNs and Vision Transformers (ViT) has shown promising results in detecting key activities from newborn resuscitation videos, but also highlighted the challenges in recognizing such fine-grained activities. This work investigates the potential of generative AI (GenAI) methods to improve activity recognition from such videos. Specifically, we explore the use of local vision-language models (VLMs), combined with large language models (LLMs), and compare them to a supervised TimeSFormer baseline. Using a simulated dataset comprising 13.26 hours of newborn resuscitation videos, we evaluate several zero-shot VLM-based strategies and fine-tuned VLMs with classification heads, including Low-Rank Adaptation (LoRA). Our results suggest that small (local) VLMs struggle with hallucinations, but when fine-tuned with LoRA, the results reach F1 score at 0.91, surpassing the TimeSformer results of 0.70.

\end{abstract}
\section{Introduction}
\label{sec:intro}

The day-of-birth carries the highest risk of death during a whole lifespan. In 2022 approximately one million newborns died world wide during the first 24 hours of life \cite{noauthor_who_nodate}, and a large percentage of these are caused by birth asphyxia, i.e. oxygen deprivation in neonates during or right after birth. Survivors after newborn resuscitation are at increased risk of life-long impairment \cite{ersdal_early_2012}. Around 10\% of newborns need some help to start breathing, and around 4-6\% need ventilation in Norway \cite{bjorland_incidence_2019}, with similar numbers expected in other countries.
Resuscitation and stabilization activities frequently include ventilation, stimulation (rubbing the torso) and suction to remove fluids, and less frequent heart compression, in accordance wtih international guidelines \cite{madar_european_2021}. Current practice demonstrates significant quality variations, with guidelines followed in only one-third of cases \cite{bjorland_compliance_2020}.

Objective and detailed documentation of stabilization and resuscitation activities is rarely collected in clinical practice, yet it could greatly support debriefing, quality improvement, and research to inform and strengthen guidelines.  Some previous work has proposed using video and computer vision based methods to automatically extract timelines of resuscitation activities. In 2020 Meinich-Bache et.al. \cite{meinich-bache_activity_2020} proposed ORAA-net a multi step approach consisting of object detection, region proposal, activity recognition, and finally activity timeline generation on a study conducted in Tanzania.   
The object detection and region proposal was done with a YOLO$_{v3}$ backbone, finetuned for objects like suction device, ventilator, hands with and without gloves, partly from actual resuscitation videos and partly from a generated synthetic dataset. The activity recognition was based on multiple i3D backbones combined with optical flow, finetuned for binary detection of activities around the relevant objects, for example detect actual ventilation from the region around a detected ventilator, as sometimes the ventilator is in the video but not used for ventilation.  The method was trained in a supervised setup. 
In 2024, Rizvi et. al. \cite{rizvi_semi-supervised_2024} proposed a semi-supervised approach for newborn resuscitation activities timeline generation.  The architecture is in one way simpler, as it is a one step approach using an adapted SVFormer \cite{xing_svformer_2023}, a vision extension to the TimeSformer \cite{bertasius_is_2021} model. In \cite{rizvi_semi-supervised_2024}, the input is the whole videoframe covering a resuscitation table, without object detection. The semi-supervised learning is proposed with a student-teacher, combining strong and weak augmentation and a joint loss function.  
Both of these works are promising, but they also show how challenging such fine-gran activity recognition is, and that there is improvement potential for this application.

In recent years, rapid advancements in generative AI (GenAI) have opened up a wide range of possibilities across domains, including healthcare, education, and creative industries.  Different GenAI models can now perform tasks such as generating realistic text with Large Language Models (LLM) \cite{chang_survey_2024}, summarizing clinical notes, interpreting medical images through vision-language models (VLM), and analyzing video content to extract meaningful insights \cite{lee_multimodal_2025}, \cite{hartsock_vision-language_2024}.  Large Multimodal Models (LMM) are built to handle multiple modalities together, and are usually built on top of an LLM. If used for vision and language, an LMM can function similarly to a VLM, like LLaVa (Large Language and Vision Assistant) \cite{liu_visual_2023}, but it can also include other modalities.


Newborn resuscitation videos are highly sensitive, and using cloud-based models impose multiple challenges: 
\textbf{i) Content filtering limitations in cloud services:}
Many cloud-based AI models incorporate automated filters designed to block violent or explicit content. However, our experience with simulated resuscitation videos shows that these filters can misclassify legitimate medical procedures, such as ventilation and stimulation, as violent content, thereby preventing accurate analysis or model training.
\textbf{ii) Privacy and compliance:}
Newborn resuscitation videos contains highly sensitive medical and personal information. To ensure compliance with privacy regulations such as GDPR, processing such data requires secure, specialized infrastructure that offers strong access control, data protection, and on-premise or compliant cloud alternatives.

In this paper we explore a combination of LLMs and VLMs for the purpose of generating activity timelines of newborn resuscitation episodes. We  compare that to the use of a fully supervised TimeSFormer \cite{bertasius_is_2021}, as was used on activity recognition in newborn resuscitation, as the baseline compared to semi-supervised learning in \cite{rizvi_semi-supervised_2024}. 
Given the aforementioned challenges, we explore the use of open-source, \emph{edge-based} VLMs and LLMs, allowing us to maintain full control over both the models and the data handling in future deployments. To mitigate privacy concerns and data availability during the exploration and model development phase, we utilize videos from simulated resuscitation events.  The simulation data are  videos recorded in a hospital, using a resuscitation table and ventilator, with a newborn manikin serving as substitute for a real infant.   

\section{Data material}
\label{sec:format}
The data used in this study is RGB video without sound, recorded at a hospital using the same equipment as used in clinical newborn resuscitation setting, but as a simulation setup.  A manikin of type "Newborn Anne" from Laerdal medical was used as substitute for a real infant.  The simulation was done with different people, some trained medical experts, and some not-medical, but working in the project, with knowledge about newborn resuscitation activities. In all sessions a trained medical professional was present, leading the session.

In total 13.26 hours of video has been collected in total over 10 different sessions. 
The different resuscitation activities were manually annotated in time using the ELAN tool\footnote{Max Planck Institute for Psycholinguistics, The Language Archive, Nijmegen, The Netherlands, https://archive.mpi.nl/tla/elan} \cite{wittenburg_elan_2006}.  No spatial annotation was done.   The annotation was done by a non-medical that was first trained by a medical expert. 
Let the size of a videoframe be denoted $h \times w$, as height $\times$ width, and the number of frames denoted $N_{f}$. From the annotated videos, 3 seconds clips are extracted, denoted ${\bf X}\in \mathcal{R}^{h\times w \times c \times N_{f}}$, where $h \times w = 768 \times 1024 $, $c=3$ for RGB color channels, and $N_{f} = 75$ for a 3 sec clip with frame rate 25.  Each ${\bf X}_{i}$ can be associated with none, one or more of the  corresponding labels, $y_{text}\in \{$ventilation, stimulation, suction, baby on table$\}$ in a multilabel setup. Ventilation and suction can not happen at the same time, but the other labels can be overlapping.  Translated to a label vector ${\bf y} \in \mathcal{R}^{cl}$, where $cl$=number of classes, here 4.  ${\bf y}_{i} = [1 \: 1 \: 0 \: 1]^{T}$ means that for video clip $X_{i}$ the newborn is visible on the table and ventilation and stimulation is being performed, but suction is not performed. ${\bf y}_{i} = [0 \: 0 \: 0 \: 1]^{T}$ simply means that the baby is on the table but none of the three main activities are being performed, and ${\bf y}_{i} = [0 \: 0 \: 0 \: 0]^{T}$ is a clip not containing a baby at the table.
For a specific label to be activated for a clip, ${\bf X}_{i}$, minimum 50\% of the 3 second interval must be annotated with that label.  The dataset $\mathcal{D}=\{({\bf X}_{i},{\bf y}_{i})\}_{i=1}^{N}$, is divided into a trainingset $\mathcal{D}_{tr}$ and a test $\mathcal{D}_{te}$ so that $\mathcal{D} = \mathcal{D}_{tr} \cup \mathcal{D}_{te}$, and $\mathcal{D}_{tr} \cap \mathcal{D}_{te} = \varnothing$.

\begin{figure}[htbp]
    \centering
    \begin{subfigure}[b]{0.23\textwidth}
        \centering
        \includegraphics[width=\linewidth]{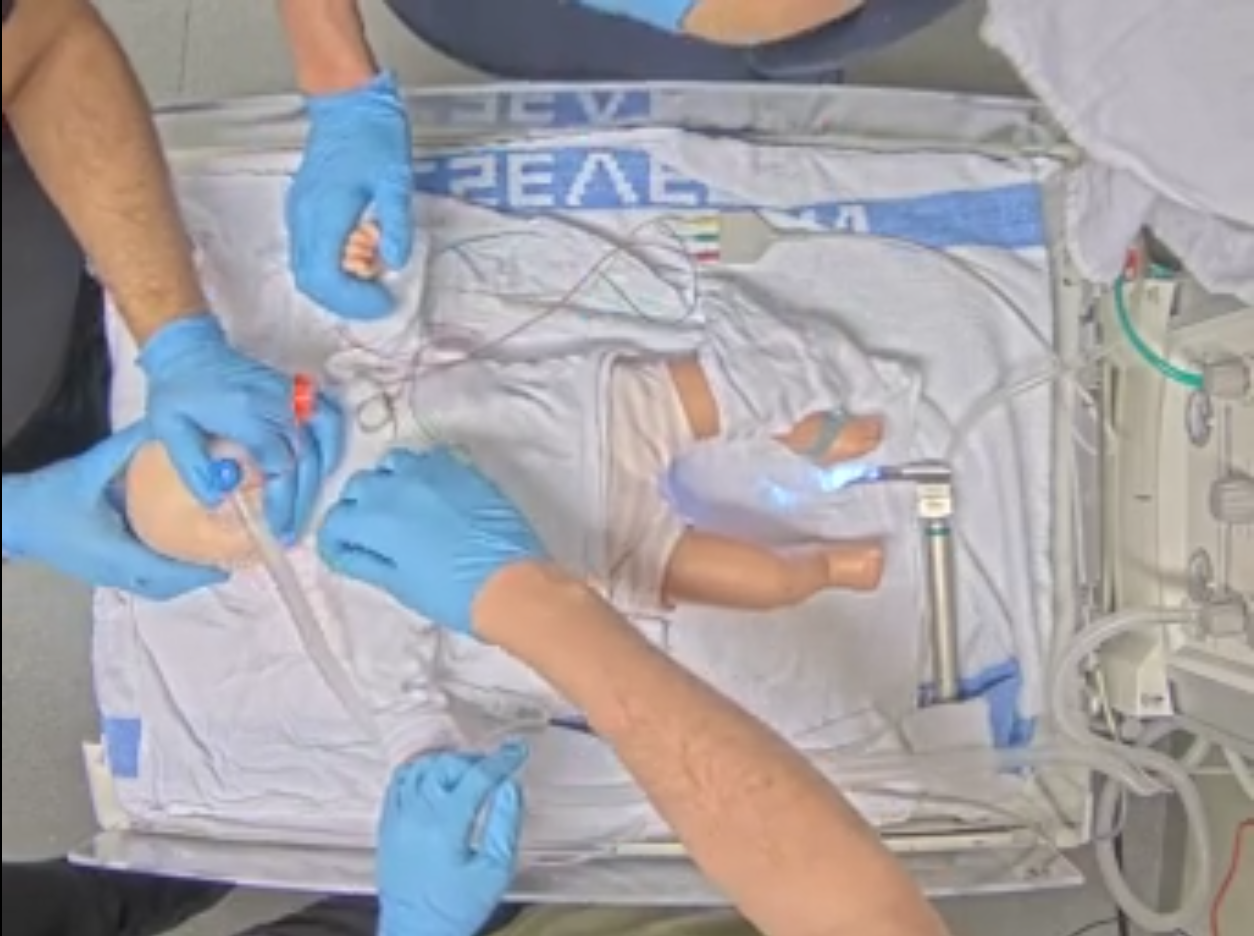}
        \caption{Ventilation}
    \end{subfigure}
    \hfill
    \begin{subfigure}[b]{0.23\textwidth}
        \centering
        \includegraphics[width=\linewidth]{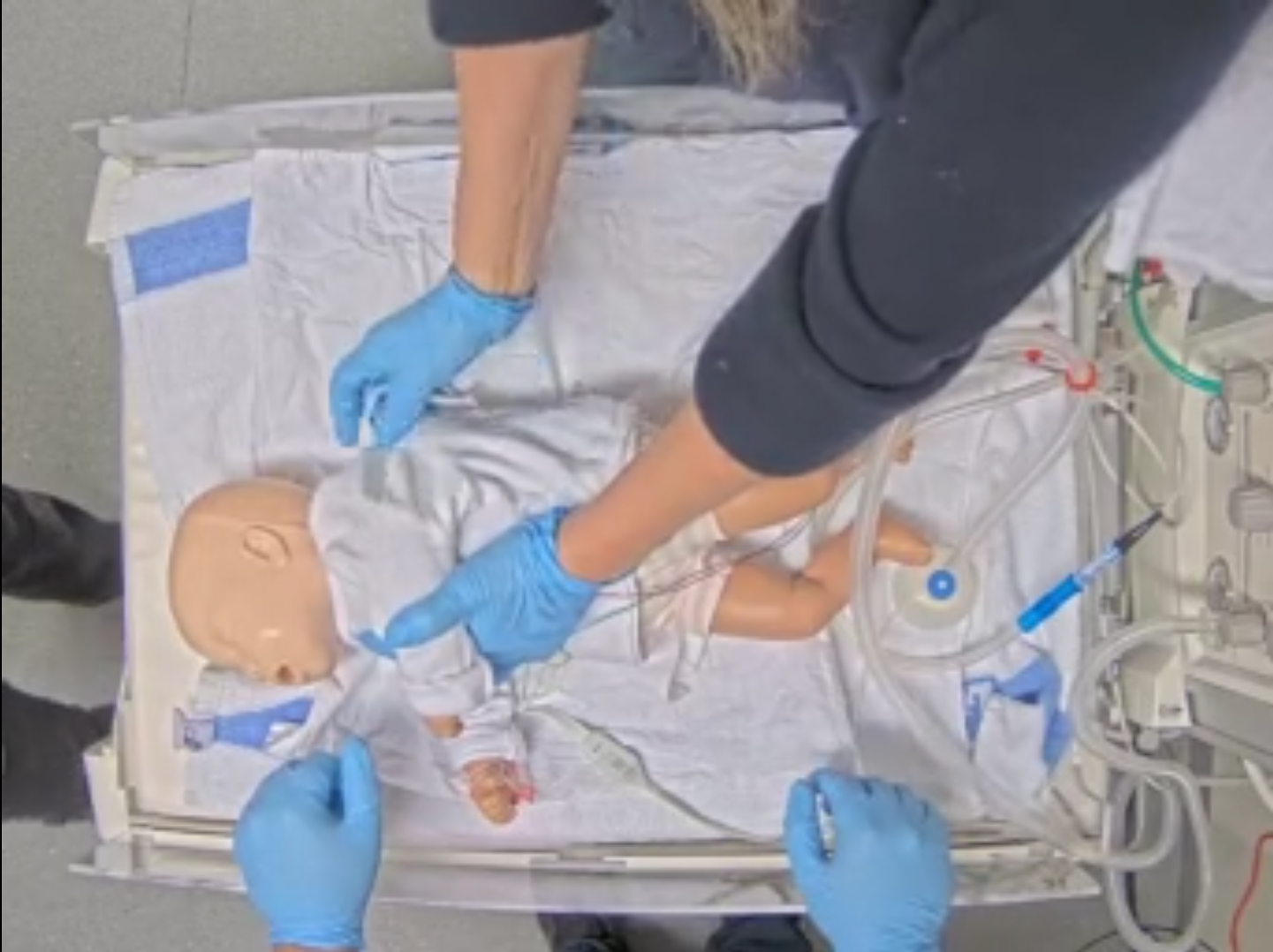}
        \caption{Stimulation}
    \end{subfigure}
    \vspace{0.5em}
    \begin{subfigure}[b]{0.23\textwidth}
        \centering
        \includegraphics[width=\linewidth]{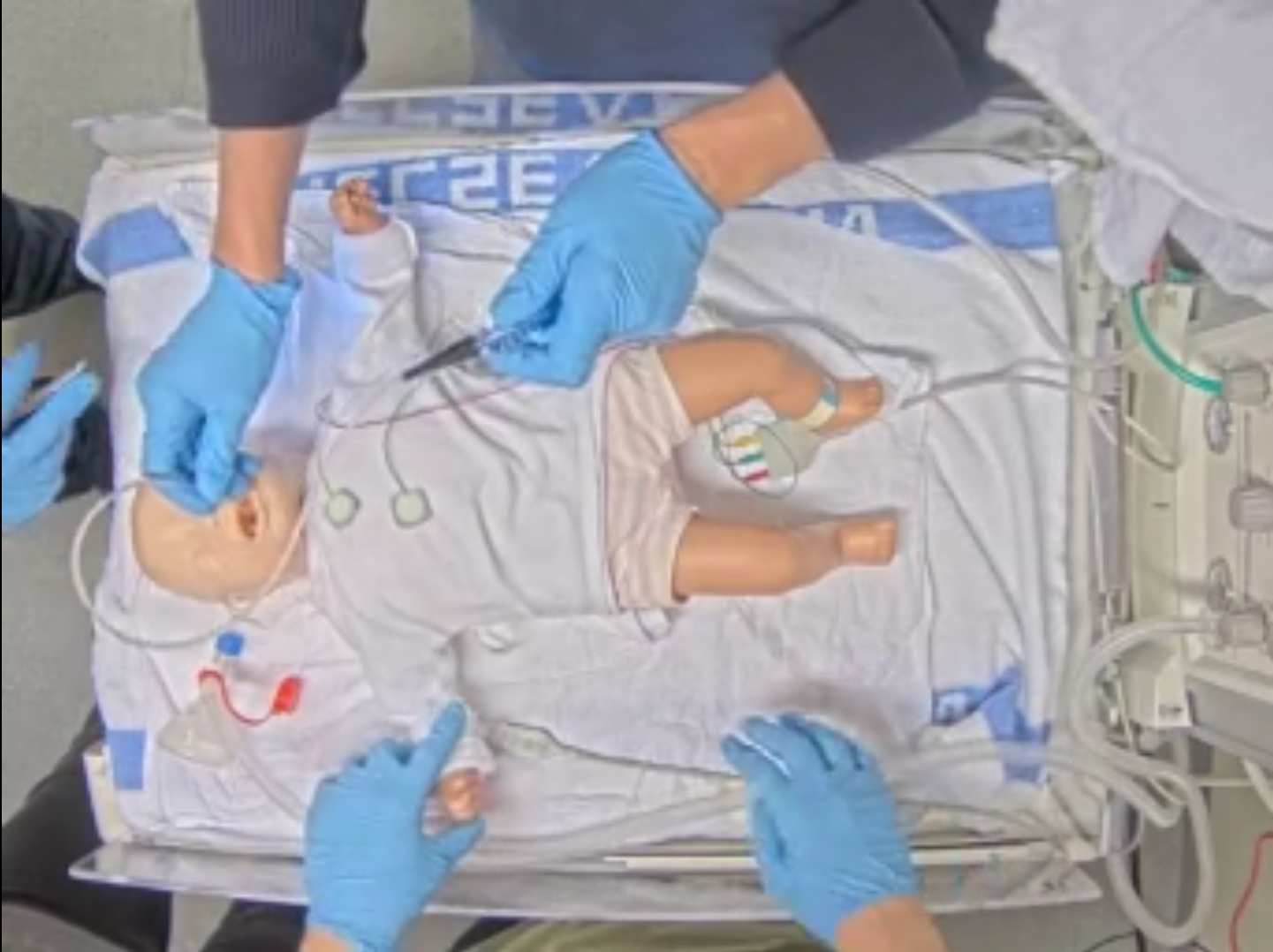}
        \caption{Suction}
    \end{subfigure}
    \hfill
    \begin{subfigure}[b]{0.23\textwidth}
        \centering
        \includegraphics[width=\linewidth]{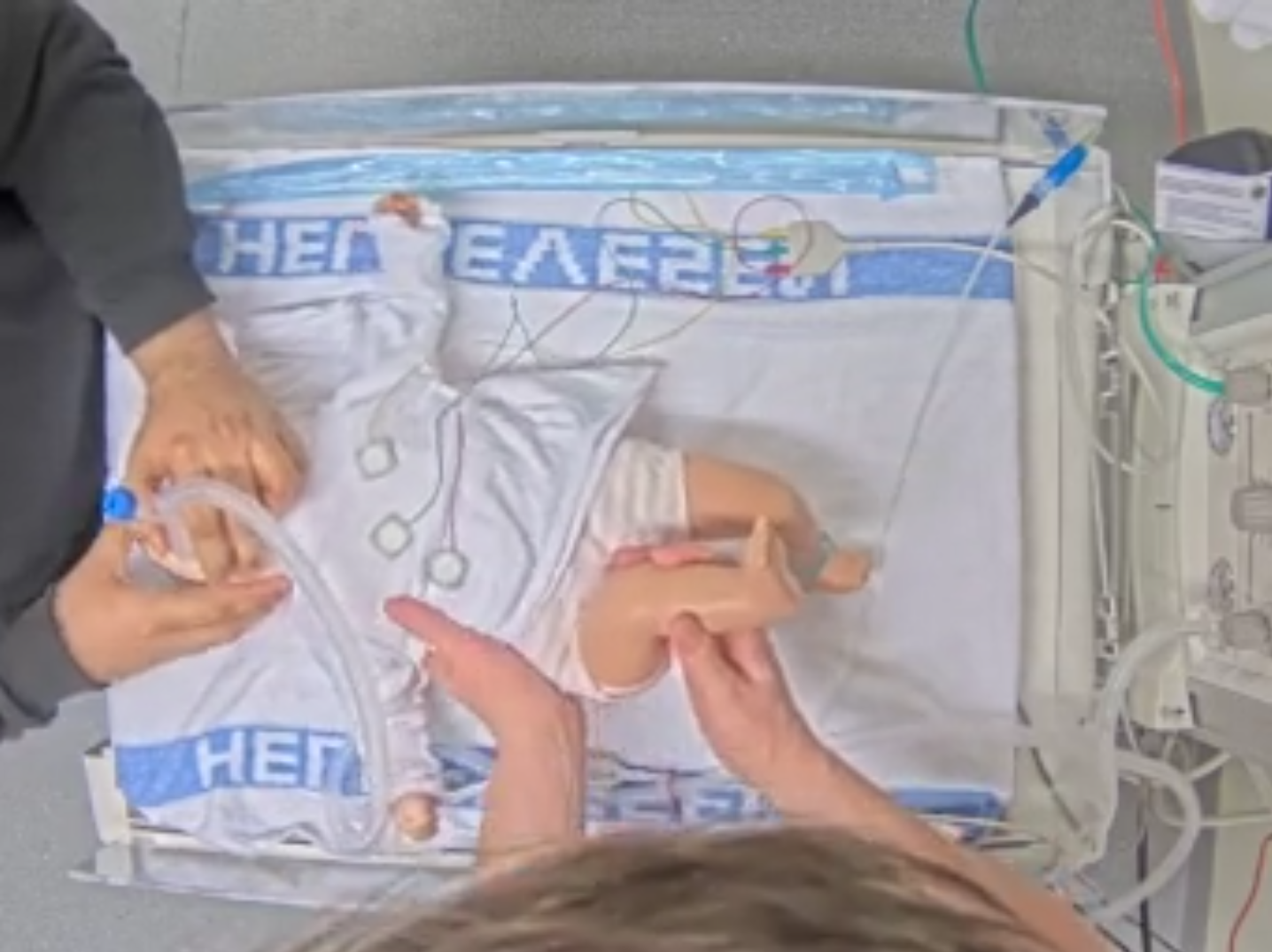}
        \caption{Ventilation and Stimulation}
    \end{subfigure}

    \caption{Examples of video frames from the different activities.}
    \label{fig:frames_exmpl}
\end{figure}

Examples of frames from the different activities are shown in Figure \ref{fig:frames_exmpl}. 

\section{Method}

This study explores and compares several approaches to multi-label classification using a Video Language Model. The open source LLaVA-NeXT Video \cite{zhang2024llavanextvideo} (7 billion parameters) is used as VLM, and Mistral 7B \cite{mistral7B} as LLM in the following, chosen for their reported benchmarks combined with the availability. In preliminary experiments other VLMs were tested, but LLaVA-Next Video was the most promising.   Different approaches are detailed in the following.

\subsection{Baseline - TimesSFormer}
We seek a predicted label vector ${\bf\hat{y}}_{i}=f_{\theta}({\bf X}_{i})$, where $f_{\theta}()$ corresponds to a model with parameters $\theta$. Often preprocessing is needed, as the videoclip ${\bf X}_{i}$ is larger than the expected video input of the models, giving ${\bf\hat{y}}_{i}=f_{\theta}(g({\bf X}_{i}))=f_{\theta}({\bf X}^{r}_{i})$, where $g({\bf X}_{i}))= {\bf X}_{i}^{r}$ can be understood as a preprocessing consisting of resampling and/or normalization. $\theta$ is learned by supervised learning using $\mathcal{D}_{tr}$.
In the case of TimeSformer \cite{bertasius_is_2021}, $f_{\theta}$ corresponds to a divided space-time attention transformer pre-trained on Something-Something V2.
The model receives a fixed-length sequence of $T\leq N_f$ RGB frames;
each frame ${\bf x}_{i,t} \in \mathbb{R}^{h \times w \times 3}$ for $t = 1, \dots, T$ is divided into a grid of $P \times P$ patches, producing $N_p = \frac{HW}{P^2}$ patches per frame.
Each patch is then flattened and mapped to a $d$-dimensional vector $\bf z_{i,t}^{(j)}=\bf E_{\text{patch}}\cdot\text{vec}(\bf x_{i,t}^{(j)})\in\mathbb{R}^d$ with $j=1,...,N_p$, ${\bf E}_{\text{patch}}\in\mathbb{R}^{d\times (P^2\cdot 3)}$ learnable patch embedding matrix, and $\bf x_{i,t}^{(j)}$ $j$-th patch of frame $t$.
The patch embeddings are augmented with spatial and temporal positional encodings: $\bf h_{i,t}^{(j)} = \bf z_{i,t}^{(j)} + \bf p_{\text{space}}^{(j)} + \bf p_{\text{time}}^{(t)}$.
The resulting tensor, of shape $(T \cdot N_p, d)$, is passed through a stack of $L$ transformers layers, with self-attention factorized across spatial tokens within a frame, and then across temporal tokens across frames: $\mathbf h_l^{out} = MSA_{\text{temp}}(MSA_{\text{space}}(\mathbf h_{l-1}^{out}))$ with $MSA$ Multi-head Self Attention \cite{bertasius_is_2021}.
Finally, a classification head aggregates the token representations and maps them to the label space via:
\begin{equation}
    \mathbf{\hat y}_i = \sigma( \mathbf W_{cls}\cdot Agg(\mathbf h_L^{out})+\mathbf b_{cls}) \in [0,1]^C
\label{eq:clas_head}
\end{equation}
with $Agg$ aggregation function using a CLS token, ${\mathbf W}_{\text{cls}} \in \mathbb{R}^{C \times d}$ classification weight matrix and $\mathbf b_{\text{cls}}$ bias term, initialized as $\mathbf b_{\text{cls}, c} = \log(p_c) - \log(1-p_c)$, with $p_c$ prior probability for class $c$.

\begin{figure}[htbp]
    \centering
    \includegraphics[width=0.5\linewidth]{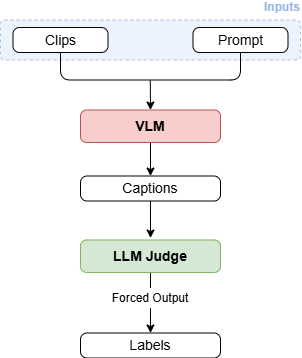}
    \caption{ZSC-J: VLM captioning + LLM judge. VLM is frozen, prompt and judge is tuned}
    \label{fig:judge}
\end{figure}
\subsection{Zero-Shot Classification with Constrained Output Format, ZSC-CO}

In the ZSC-CO approach, a zero-shot inference is found by prompting the VLM directly with an instruction designed to elicit a structured output. We now aim to predict a label vector $\hat{\mathbf{y}}_{i} = f_{\gamma}(g(\mathbf{X}_i)| p,{temp})$. Where $p$ is a textual prompt encoding the set of candidate labels and task-specific instructions, where dependencies between classes are explicit. ${temp}$ is the temperature.  $f_\gamma$ is the VLM, here LLaVA-Next Video with parameters $\gamma$, which jointly processes the visual input $\mathbf{X}^r_i$ and the prompt $p$ to produce the predicted label vector $\hat{\mathbf{y}}_i$.  The first output is a comma-separated list of the actions present in the clip.  $f_{\gamma}$ is kept frozen, $\mathcal{D}_{tr}$ is used for prompt design.
The prompt design consists in finding a balance between the additional medical information to give to a model and not to make it hallucinate. 
By providing the model with too detailed information in the prompt about events (e.g. describing how the ventilation mask looks) the model more often hallucinated.

    
    


The approaches that include the LLaVA-NeXT Video model capture temporal features in a less direct way than the baseline (TimeSformer), since the Vision Tower of the VLM is a frozen ViT-based encoder.
The encoder captures spatial features using the same patch embedding scheme as TimeSFormer. However, unlike TimeSFormer’s bi-directional temporal modeling, the visual tokens generated by the vision tower are passed to the LLM in chronological order and processed causally. Specifically, a masking mechanism ensures that each frame is conditioned only on preceding frames, with no access to future ones. This causal processing during pre-training enables the model to learn temporal dependencies through frame progression. 


\subsection{Zero-shot Binary Testing, ZS-B}
In the prompt based zero-shot binary classification setup, the VLM is asked separate yes/no question for each class. The objective here is to try to limit the hallucinations we experience from the multi-label approach, by simplifying the description and prompt to the VLM.
The prompts are event-specific and this gives us the possibility to be more precise in the context description and tailor each question to the characteristics of the target event.

\begin{figure}[htbp]
    \centering
    \includegraphics[width=0.6\linewidth]{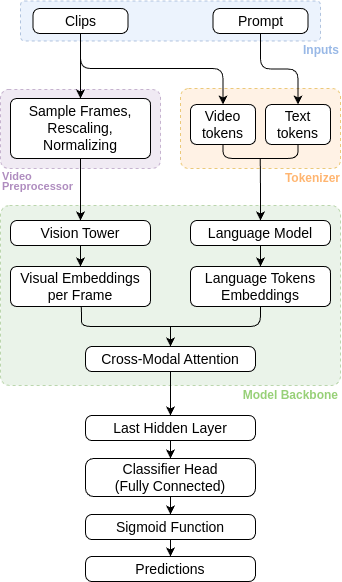}
    \caption{Architecture of the Fine-Tuned Model: the fused representation of Prompt and Video is passed through the Classification Head. FT-LC has trainable \textit{Classifier Head}, FT-C-LoRA have in addition trainable parameters within the \textit{Cross-Modal Attention} block.}
    \label{fig:VLM}
\end{figure}

\subsection{Zero-Shot Captioning with LLM Judge, ZSC-J}
The ZSC-J method aims to reduce the hallucinations of the VLM model distributing the context understanding between two different models. The VLM is simply asked to caption the clip in a zero-shot setup, with an LLM acting as a judge, that maps the caption to a binary vector over the defined label space, as illustrated in Figure \ref{fig:judge}.
Let $j$ represent the judge prompt, now we seek $\hat{\mathbf{y}}_{i} = f_{\gamma}(g(\mathbf{X}_i)| p,j,{temp})$.  $f_{\gamma}$ is frozen, $\mathcal{D}_{tr}$ is used for prompt and jugde design.
Specifically, the resulting VLM prompt asks the model to caption the clip:
\begin{shaded}
\ninept
    You are in a simulation of a newborn resuscitation scenario.
    You will be shown a short video clip of a newborn resuscitation simulation.
    You need to caption the video, describing exactly:
    - Who is present in the video
    - What actions are being performed
    - What objects are being used.
    Be explicit and precise in your description, focus on medical details.
\end{shaded}

The LLM judge performs a binary text classification task: given the caption, it classifies each clip 4 times and begins each answer with 0 or 1, allowing us to easily record the scores.

\subsection{Fine Tuning the VLM with a Classifier Head }
The core method of the study consists in the supervised fine-tuning of the VLM for multi-label classification, achieved by appending a classification head to the model's final token representations.
The VLM processes video and prompt through the full backbone: the video is first passed through the vision tower, and the prompt is tokenized via the text tokenizer. 
These two streams are then fused via cross-modal attention layers, producing a sequence of hidden states.
From the final hidden layer, we extract a designated token to serve as the aggregate video representation, which is then passed to the classification head, as it is shown in figure \ref{fig:VLM}.

The classification head has the same structure as the head we put on the TimeSformer (equation \ref{eq:clas_head}), with a key difference.
Instead of using the CLS token embedding, we compute the mean of the video token embeddings.
Since the textual prompt remains identical for all clips, using the CLS token would lead to similar feature representations.
Averaging only the video token embeddings ensures that the classifier focuses on the visual content of each clip;
nevertheless, this still allows the prompt to influence the embeddings through cross-modal attention during the transformer processing.
For training, a weighted binary cross entropy loss has been used, due to the high class imbalance and the nature of the task:
\begin{equation}
    \mathcal{L}_{\text{WBCE}} = - (w_+\cdot y \cdot \log(\hat y) + (1-y) \cdot \log(1-\hat y))
\end{equation}
with $\hat y$ predicted probability, $y$ true binary label and $w_+$ weights for the positive class.
\begin{figure}
    \centering
    \begin{subfigure}[b]{0.23\textwidth}
        \centering
        \includegraphics[width=0.9\linewidth]{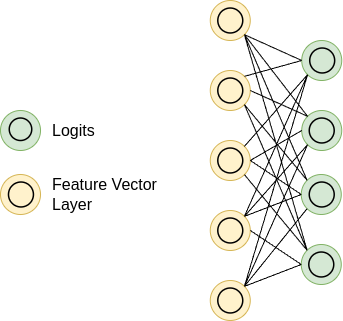}
        \caption{Classification Head, from 4096 hidden size to 4 logits}
        \label{fig:Clas_head}
    \end{subfigure}
    \hfill
    \begin{subfigure}[b]{0.23\textwidth}
        \centering
        \includegraphics[width=\linewidth]{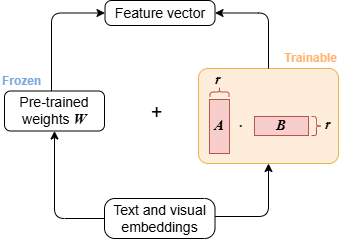}
        \caption{Low Rank Adaption within each Cross Modal Attention layers}
        \label{fig:LoRA}
    \end{subfigure}
    \caption{Trainable Parameters in the FT-LC (a) and FT-C-LoRA (a) and (b) models.}
    \label{fig:enter-label}
\end{figure}
Two training approaches has been used: \\

\textbf{Linear classifier head only, FT-LC}: 
We freeze the VLM backbone and train only the final classifier head (Figure \ref{fig:Clas_head}). This approach allows us to store feature vectors and efficiently run multiple hyperparameter-tuning experiments. Hyperparameter tuning covered both algorithmic and architectural parameters.\\

\textbf{Classifier head + LoRA , FT-C-LoRA}: Low-Rank Adaption (LoRA) modules, as described in \cite{hu2022lora}, are injected into the VLM's cross attention layers, allowing partial fine-tuning of the backbone with a minimal number of training parameters.
    Specifically, given the weight matrix $W_0\in\mathbb{R}^{d\times k}$, LoRA adds a trainable low-rank update in the form of two matrices $A$ and $B$, such that the adapted weight become $W = W_0+ BA$.
    The backbone (figure \ref{fig:LoRA}) is trained with a smaller learning rate than the classifier (figure \ref{fig:Clas_head}) in order to avoid catastrophic forgetting and overfitting.
    Specifically, for LoRA setting we followed the instruction tuning provided by Liu et al. 
    \cite{liu2023visual},
   \cite{liu2024improved} 

\section{Experiments and results}
The dimension of the dataset is $X \in \mathcal{R}^{....x3x75}$, however, the different models require different input sizes, thus a preprocessing is done resampling $X_{i}$ to $X^r_i$. 
Specifically, the TimeSformer model, pre-trained on SSV2, accepts 8 frames videos, while LLaVA-NeXT Video accepts 16 frames videos. 

The multilabel classification performed by the models involves recognizing fine-grained details (e.g., whether the suction tube is inserted into the newborn's mouth/nose, whether a tube connects to a ventilator mask or suction device, and whether hand movements involve multiple health care providers or single-hand stimulation), and the choice of evaluation metrics is important.
Due to the strong class imbalance and the interdependence of the labels, F1 score is considered a good choise. 
For the hyper-parameter optimization, the macro F1 score (average of the F1 score per class) has been used as indicator.

The optimization has been performed with an Optuna setup, using the Tree-structured Parzen Estimator algorithm \cite{bergstra2011algorithms}, supported by pruners implemented to stop unpromising trials early.
Specifically, that algorithm has been used for tuning the TimeSformer model and the FT-LC. 
This because the optimization process is highly time-consuming and resource-intensive, and it would have been unaffordable to run over the entire VLM multiple times. 
Instead, for the classification head fine tuning, the VLM has been run once, saving all the feature vectors and allowing fast and efficient experiments.
The optimization results obtained for FT-LC were reused for training FT-C-LoRA, as both models share the same architecture. 
Due to the high computational cost of training FT-C-LoRA, running separate optimization experiments on it was not feasible.
Fine-tuning is done using the trainingset, ${\cal D}_{tr}$. Results on the test set ${\cal D}_{te}$ are seen in Table\ref{tab:results}.
The Baby Visible prediction is missing from ZSC-CO results due to model hallucinations when forced to focus on all 4 classes. We excluded this basic class and focused on the most interesting ones to generate reasonable descriptions without hallucination.

\begin{table}[]
    \centering
    \begin{tabular}{c|c|c|c|c|c}
         Methods & B.V. & Vent. & Stim. & Suct. & mAv \\
         \hline
         TimeSformer & 0.99 & 0.51 & 0.71 & 0.57 & 0.70 \\
         ZSC-CO & / & 0.45 & 0.24 & 0.22 & 0.23 \\
         ZS-B & 0.99 & 0.46 & 0.50 & 0.19 & 0.54 \\
         ZSC-J & 0.92 & 0.40 & 0.36 & 0.11 & 0.45\\
         FT-LC & 0.99 & 0.60 & 0.68 & 0.19 & 0.62 \\
         \textbf{FT-C-LoRA} & \textbf{0.996} & \textbf{0.92} & \textbf{0.79} & \textbf{0.94} & \textbf{0.91} \\
       
    \end{tabular}
    \caption{F1-score per class for each method on the test set.   B.V., Vent, Stim and Suct means baby visible, ventilation, stimulation, and suction. mAv is the macro average F1 score}
    \label{tab:results}
\end{table}
       
 
\section{Discussion}
  The dataset contains highly fine-grained activities, with video clips that are visually similar and differ only in subtle ways. The background and camera angle remain constant across all clips, and each scene consistently features two to three individuals wearing blue gloves interacting with the same manikin. The distinguishing features lie in small, nuanced details of the actions, making the classification task considerably more challenging than many standard activity recognition benchmarks in the literature.
  

  The TimeSFormer results are somewhat lower than in the paper by Rizvi et.el. \cite{rizvi_semi-supervised_2024}, but this can be due to the different datasets.  The zero-shot models, ZSC-CO, ZS-B, and ZSC-J all fall short of the TimeSFormer baseline.  Our experiments are constrained to relatively small models due to the need to perform both training and inference locally, driven by privacy concerns related to sensitive video content and the limitations of content filtering in cloud services. A recurring issue with the zero-shot approaches is hallucination, which may be mitigated by larger models but could also reflect the inherent difficulty of the fine-grained task itself. In contrast, fine-tuning significantly improves performance. Among the fine-tuned models, FT-C-LoRA outperforms FT-LC, probably influenced by its larger number of learnable parameters. We attempted to increase the capacity of FT-LC by using deeper classifier heads, but this led to rapid overfitting. The best results are achieved with the FT-C-LoRA model, which attains the highest individual F1 scores and the best macro-average F1 score of 0.91.

  
  
\section{Conclusion}
We addressed a challenging fine-grained activity recognition task from newborn resuscitation videos, and asked \textit{Can local vision-language models improve fine-grained activity recognition over vision transformers?} To explore this, we compared the ViT-based TimeSFormer model with LLaVA-NeXT Video as the vision-language backbone and Mistral 7B  as the language model.  
We could not make the zero-shot models perform in pairs with the TimeSFormer result of mAv F1 score 0.7.  However, when we fine-tuned a classification head, keeping the rest of the LLaVa-NeXT Video frozen, we observed a substantial performance gain. The best results were achieved using a LoRA-based classifier head, FT-C-LoRA, reaching a mAv F1 score of 0.91.
These findings highlight the potential of local VLMs for complex, privacy-sensitive video analysis tasks. However, they also suggest that task-specific fine-tuning of the classification head is essential to realize this potential in practice.



\ninept
\bibliographystyle{IEEEbib}
\bibliography{icip2025_refs}

\end{document}